\def\year{2020}\relax
\documentclass[letterpaper]{article} 
\usepackage{aaai20}  
\usepackage{times}  
\usepackage{helvet} 
\usepackage{courier}  
\usepackage[hyphens]{url}  
\usepackage{graphicx} 
\usepackage{graphicx}  
\usepackage{amsmath,amssymb}
\usepackage{booktabs} 
\usepackage{bm}
\usepackage{url}
\usepackage{helvet}
\usepackage{courier}
\usepackage{graphicx}
\usepackage{multirow}
\graphicspath{{./}{./Fig/}{./Figs/}}
\usepackage{algorithm}
\usepackage{algorithmic}
\usepackage{enumitem}
\usepackage{newunicodechar}
\newunicodechar{ﬁ}{fi}

\title{A Proposal-based Approach for Activity Image-to-Video Retrieval}

\author{Ruicong Xu, Li Niu,\thanks{Corresponding author} Jianfu Zhang, Liqing Zhang\footnotemark[1]\\
MoE Key Lab of Artificial Intelligence, \\
Department of Computer Science and Engineering, \\
Shanghai Jiao Tong University, Shanghai, China. \\
\{ranranxu, utscnewly, c.sis\}@sjtu.edu.cn, zhang-lq@cs.sjtu.edu.cn}

\begin{document}
\maketitle
\begin{abstract}
Activity image-to-video retrieval task aims to retrieve videos containing the similar activity as the query image, which is a challenging task because videos generally have many background segments irrelevant to the activity. In this paper, we utilize R-C3D model to represent a video by a bag of activity proposals, which can filter out background segments to some extent. However, there are still noisy proposals in each bag. Thus, we propose an Activity Proposal-based Image-to-Video Retrieval (APIVR) approach, which incorporates multi-instance learning into cross-modal retrieval framework to address the proposal noise issue. Specifically, we propose a Graph Multi-Instance Learning (GMIL) module with graph convolutional layer, and integrate this module with classification loss, adversarial loss, and triplet loss in our cross-modal retrieval framework. Moreover, we propose geometry-aware triplet loss based on point-to-subspace distance to preserve the structural information of activity proposals. Extensive experiments on three widely-used datasets verify the effectiveness of our approach.
\end{abstract}

\section{Introduction}
Cross-modal retrieval task has attracted considerable research attention in the field of retrieval task. With the rapid development of video applications, a specific type of retrieval task, Activity Image-to-Video Retrieval (AIVR), comes into our sight.
The goal of AIVR task is to retrieve the videos containing the similar activity as the image query, which expands its value in widespread applications. One daily-life example is news videos searching with a provided photo containing a particular activity. Another example is fitness videos recommendation based on a sports picture.

The key idea of cross-modal retrieval is to learn a common feature space, where cross-modal data of relevant semantic can be close to each other. Although there are abundant methods for cross-modal retrieval like text-image retrieval~\cite{FengWL14,HardoonSS04,PengHQ16,WangHWWT16,WangHWWT13}, few methods~\cite{AraujoG18,XuYS0S17} are proposed for image-video retrieval. However, these methods are not specifically designed for AIVR task. 
AIVR task is in high demand of meaningful video representations, because a video may contain background segments irrelevant to the activity and poor video representations without considering noisy background segments will lead to inferior performance of AIVR task.

Recently,  RNN~\cite{NgHVVMT15,SrivastavaMS15} and 3D CNN~\cite{JiXYY13,TranBFTP15,QiuYM17} are used to extract deep learning-based video representations.
As an extension of 3D CNN, R-C3D \cite{XuDS17} can generate candidate temporal regions containing activities and filter out noisy background segments to obtain the superior activity video representations. Therefore, we take advantage of R-C3D model to generate temporal proposals that are most likely to contain activities and extract one feature vector for each proposal, leading to a bag of proposal features for each video. This paper is the first to target at AIVR task by utilizing activity proposals for videos. 

\begin{figure*}[tp]
	\begin{center}
		\includegraphics[width=1.0\linewidth]{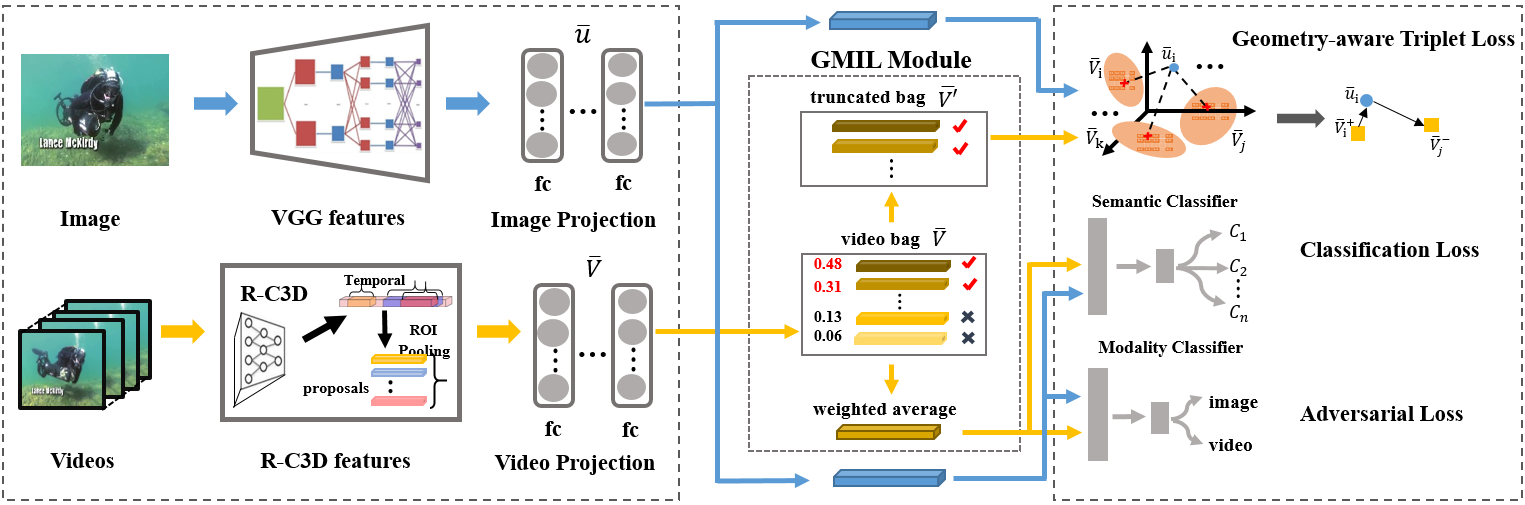}
	\end{center}
	\caption{The flowchart of our proposed approach. The image features and bags of activity proposal features for videos are extracted by VGG~\cite{SimonyanZ14a} and R-C3D~\cite{XuDS17} models respectively, and then projected into a common feature space. Our retrieval framework consists of triplet loss, classification loss, and adversarial loss. We incorporate Graph Multi-Instance Learning (GMIL) module into retrieval framework to address the proposal noise issue. We also design geometry-aware triplet loss based on truncated bag of activity proposals. Best viewed in color.}
	\label{fig1}
\end{figure*}

In this paper, we propose an Activity Proposal-based Image-to-Video Retrieval (APIVR) approach for AIVR task. The major innovation in our paper is incorporating Graph Multi-Instance Learning (GMIL) module into cross-modal retrieval framework to address the proposal noise issue. As illustrated in Figure~\ref{fig1}, our cross-modal retrieval framework is based on Adversarial Cross-Modal Retrieval (ACMR) proposed in \cite{WangYXHS17}, in which image features and activity proposal-based video features are projected into a common feature space steered by triplet loss, classification loss, and adversarial loss. 
To address the proposal noise issue, we treat each video as a bag and the activity proposals in each bag as multiple instances, which coincides with multi-instance learning (MIL) paradigm~\cite{IlseTW18}. We assume that there is at least one clean instance in each bag, and employ self-attention mechanism to learn different weights for multiple instances, with higher weights indicating clean activity proposals. To further consider the relation among multiple instances in each bag, we insert graph convolutional layer into MIL module, yielding a novel Graph MIL (GMIL) module.

After learning weights based on our GMIL module, we use weighted average of activity proposal features in each bag as input for the classification loss and adversarial loss in cross-modal retrieval framework, to suppress noisy activity proposals. For the remaining triplet loss, 
we propose a novel geometry-aware triplet loss, which calculates the point-to-subspace distance between image and bag of activity proposals. Considering that the noisy activity proposals may mislead the point-to-subspace distance, we use truncated bag of activity proposals based on the weights learnt by our GMIL module.
Thus, our geometry-aware triplet loss can mitigate the proposal noise issue and simultaneously preserve the geometry property of activity proposals.

The contributions of our paper are summarized as follows:
\begin{itemize}
	\item This work is the first activity proposal-based approach for activity image-to-video retrieval task. Our major contribution is incorporating multi-instance learning into cross-modal retrieval framework to address the proposal noise issue.
	\item Our two minor contributions are Graph Multi-Instance Learning (GMIL) module with graph convolutional layer and geometry-aware triplet loss based on truncated bag of activity proposals.
	\item Experiment results on three datasets, \emph{i.e.}, action-based THUMOS'14 and ActivityNet datasets, event-based MED2017 Event dataset, demonstrate the superiority of our approach compared to state-of-the-art methods.
\end{itemize}


\section{Related Work}
In this section, we provide a brief overview of video representation, cross-modal retrieval, and multi-instance learning.

\noindent\textbf{Video representations:}
Video representations play a crucial role in image-to-video retrieval task.
Recently, deep learning-based models, \emph{e.g.}, RNN~\cite{JiangWWXC18} and 3D CNN~\cite{QiuYM17}, are proposed to fully exploit spatio-temporal information across consecutive frames.
As an advanced 3D CNN model, R-C3D~\cite{XuDS17} can generate activity proposals across temporal dimension to filter out noisy background segments. Hence, we adopt R-C3D to generate video representations, which significantly facilitates the AIVR task.

\noindent\textbf{Cross-modal retrieval methods:}
Cross-modal retrieval methods fall into two major categories: binary-value based methods~\cite{YuKGC14,LinSH14,YeLTXW17,DingGZ14,XuYS0S17}  and real-value based retrieval methods~\cite{ZhaiPX14,WangHWWT16,PengHQ16,PengQHY18,WangLL16,WangYXHS17,Zhen_2019_CVPR}. 
Our cross-modal retrieval framework is based on ACMR~\cite{WangYXHS17}, which consists of classification loss, triplet loss, and adversarial loss. Our contribution is incorporating graph multi-instance learning module into cross-modal retrieval framework together with geometry-aware triplet loss.

\noindent\textbf{Multi-instance learning:}
Multi-instance learning (MIL) groups training samples into multi-instance bags, in which each bag contains at least one positive instance. Some early methods~\cite{LiKTZ09} treat one bag as an entirety or infers instance labels within each bag. Recently, deep multi-instance learning methods~\cite{ZhuLVX17,PappasP14,IlseTW18} employ pooling operators or trainable operators to aggregate multiple instances in each bag. Moreover, several graph MIL~\cite{abs-1906-04881,GuoY13} methods are proposed to exploit the graph structure of training bags in different ways, but their methods cannot be easily integrated into our cross-modal retrieval framework. 

\section{Methodology}
In this section, we introduce our activity proposal-based image-to-video retrieval approach.
\subsection{Problem Definition}

For concise mathematical expression, we denote a matrix (\emph{e.g.}, $\textbf{A}$) and vector (\emph{e.g.}, $\textbf{a}$) using an uppercase and lowercase letter in boldface respectively, and denote a scalar (\emph{e.g.}, $a$) using a lowercase letter. We use $\textbf{I}_k$ and $\textbf{A}^T$ to denote an identity matrix with size $k$ and the transpose of \textbf{A} respectively. 
By using $\textbf{vec}(\cdot)$, we perform column-wise concatenation to transform a matrix into a column vector. Moreover, we use $\left \langle\bm{x}, \bm{y}\right \rangle$ to denote the inner product of $\bm{x}$ and $\bm{y}$. 

In the AIVR task, our training process is based on mini-batches of video-image pairs $\left \{ (\bm{V}_i,\bm{u}_i)|_{i=1}^{n} \right \}$, in which $(\bm{V}_i,\bm{u}_i)$ is a pair of video and image with the same category label, and $n$ is the number of pairs in a mini-batch. Specifically, $\bm{V}_i=\left \{\bm{h}_1,\bm{h}_2,...,\bm{h}_k \right \}$ with $\bm{h}_k\in \mathbb{R}^{d_1\times 1}$ is a bag of  proposal features in the $i$-th video and $\bm{u}_i \in \mathbb{R}^{d_2\times 1}$ is the feature of the $i$-th image. Note that the dimensionalities of the image feature and activity proposal features are not equal in our problem, \emph{i.e.}, $d_1 \neq d_2$. Each pair $(\bm{V}_i,\bm{u}_i)$ is associated with a one-hot label vector $\bm{y}_i$ with the entry corresponding to its category as one. In the testing stage, given an image query, the goal of the AIVR task is to retrieve the videos related to the activity in the image. 

\subsection{Activity Proposal-based Image-to-Video Retrieval (APIVR) Approach}
As mentioned above, we represent each video as a bag of proposal features $\bm{V}=\left \{\bm{ h}_1,\bm{h}_2,...,\bm{h}_k \right \}$ and each image as a feature vector $\bm{u}$. Considering the different statistical properties and data distributions of videos and images, we project video and image features into a common feature space with the mapping function $f_{v}(\cdot)$ and $f_{u}(\cdot)$ respectively. The mapping functions are defined as
\begin{eqnarray}
f_{v}(\bm{V})\!\!\!\!\!\!\!\!\!&&=\left \{ f_{v}(\bm{h}_1),f_{v}(\bm{h}_2),...,f_{v}(\bm{h}_k) \right \}\nonumber\\
&&=\left \{ \bm{\bar{h}}_1,\bm{\bar{h}}_2,...,\bm{\bar{h}}_k \right \}=\bm{\bar{V}},\\
f_u(\bm{u})\!\!\!\!\!\!\!\!\!&&=\bm{\bar{u}},
\end{eqnarray}
where $f_v:\mathbb{R}^{d_1\times k}\rightarrow \mathbb{R}^{r\times k}$, $f_u:\mathbb{R}^{d_2\times 1}\rightarrow \mathbb{R}^{r\times 1}$. The mapping functions $f_v(\cdot)$ (\emph{resp.}, $f_u(\cdot)$) are
three fully-connected layers with model parameters denoted as $\bm{\theta}_p$.

Based on the projected features $\bm{\bar{V}}$ and $\bm{\bar{u}}$, following ACMR~\cite{WangYXHS17}, we employ three types of losses: triplet loss, classification loss, and adversarial loss. Concretely, triplet loss pulls an image close to the videos of the same category while pushing it far away from the videos of a different category. The classification loss targets at successfully separating the training samples from different categories regardless of modalities, which can preserve semantic information and simultaneously minimize the modality gap. The adversarial loss is involved in a minimax game by discriminating two modalities with a modality classifier and generating modality-agnostic representations to confuse the modality classifier, which can further reduce the modality gap. In summary, the above three types of losses jointly contribute to modality consistency and semantic distinguishability in the common feature space.

\subsubsection{Graph Multi-Instance Learning Module} \label{sec:GMIL}
In the common feature space, although we use R-C3D model to generate activity proposals from each video which are very likely to contain the activity, there still remain some noisy activity proposals irrelevant to the activity. Hence, each video is comprised of a mixture of clean and noisy proposals. If we utilize these noisy activity proposals based on the video label, the quality of semantic learning will be greatly degraded. In fact, this problem can be formulated as multi-instance learning, in which each video is treated as a bag and the activity proposals in each bag are treated as instances. Based on the assumption that there should be at least one clean instance in each bag, we expect to assign higher weights on the clean instances and lower weights on the noisy ones, so that the clean instances will play a dominant role in video bags. 

Given a bag of instances $\bm{\bar{V}}=\left \{\bm{ \bar{h}}_1,\bm{\bar{h}}_2,...,\bm{\bar{h}}_k \right \}$, inspired by \cite{IlseTW18}, we employ self-attention mechanism to learn different weights for different instances in each bag as (\ref{eqn:mil1}).
In particular, we apply a fully-connected layer $\bm{L}_1 \in \mathbb{R}^{r\times r'}$ with non-linear operation $tanh(\cdot)$ to $\bm{\bar{V}}$, producing $tanh(\bm{\bar{V}}^{T}\bm{L}_1) $. Then, 
we apply another fully-connected layer $\bm{L}_2 \in \mathbb{R}^{r'\times1}$ followed by softmax layer to obtain the $k$-dim weight vector $\bm{a}$ for $\bm{\bar{V}}$.
\begin{equation}\label{eqn:mil1}
%
\bm{a} = softmax(tanh(\bm{\bar{V}}^{T}\bm{L}_1)\, \bm{L}_2).
\end{equation}

However, the above process ignores the relation among multiple instances in each bag. To take such relation into consideration, we insert graph convolutional layer~\cite{KipfW17} into (\ref{eqn:mil1}), which can leverage the graph structure of each bag. Graph convolutional layer~\cite{KipfW17} is originally proposed for semi-supervised learning and now we employ it for multi-instance learning. Following~\cite{KipfW17}, we calculate the similarity graph $\bm{S}$ for each bag $\bm{V}=\left \{\bm{ h}_1,\bm{h}_2,...,\bm{h}_k \right \}$ during preprocessing, in which $S_{ij}$ is the cosine similarity between $\bm{h}_i$ and $\bm{h}_j$. Besides, we define $\bm{S}' = \bm{S} + \bm{I}_k$ and a diagonal matrix $\bm{D}$ with $D_{ii} = {\sum }_{j} S'_{ij}$. Then, graph convolutional layer can be represented by a $1\times1$ convolution layer with parameters 
$\bm{ \bar{S }} = \bm{D}^{-1/2}\bm{S}'\bm{D}^{-1/2}$. We insert two graph convolutional layers into (\ref{eqn:mil1}) and arrive at
\begin{equation}\label{eqn:a}
\hat{\bm{a}}= softmax(\bm{ \bar{S}} \: tanh(\bm{\bar{S}} \bm{\bar{V}}^{T}\bm{L}_1)\, \bm{L}_2).
\end{equation}
The generated $\hat{\bm{a}}$ is expected to be smoother than $\bm{a}$, \emph{i.e.}, the weights of two instances in a bag should be close when these two instances are similar. The theoretical proof and more details can be found in~\cite{KipfW17}. 

At last, we obtain the weighted average of instance features as the bag-level feature $Z(\bm{\bar{V}})=\sum_{j=1}^{k}\hat{a}_j {\bm{\bar{h}}}_{j}$. By assigning different weights on different activity proposals, we aim to focus more on the clean proposals and obtain discriminative video features.

\subsubsection{Geometry-aware Triplet Loss with GMIL}
We use triplet loss to preserve the semantic relevance of similar training samples across different modalities.
As defined in~\cite{SchroffKP15}, triplet loss is based on an anchor sample $\bm{x}$, a positive sample $\bm{p}$, and a negative sample $\bm{n}$, where $\bm{x}$ has the same category label as $\bm{p}$ yet a different category label from $\bm{n}$. 
Given a triplet $(\bm{x},\bm{p},\bm{n})$, triplet loss is used to enforce the distance between $\bm{x}$ and $\bm{p}$ to be smaller than that between $\bm{x}$ and $\bm{n}$ by a margin. 

Since our objective is to retrieve videos by a given image query, anchor sample $\bm{x}$ is an image while positive sample $\bm{p}$ and negative sample $\bm{n}$ are videos. In a mini-batch of video-image pairs $\{(\bm{\bar{V}}_i, \bm{\bar{u}}_i)|_{i=1}^n\}$, with each image $\bm{\bar{u}}_i$ being an anchor sample, we use its paired video sample as the positive sample $\bm{\bar{V}}_i^+$ and one video from a different category as the negative sample $\bm{\bar{V}}_j^-$, leading to in total $n$ triplets in a mini-batch. Then our triplet loss is formulated as
\begin{equation} \label{eqn:stl_triplet}
L_{triplet}=\sum_{i,j}\left |\: d(\bm{\bar{u}}_i,\bm{\bar{V}}_i^+)-d(\bm{\bar{u}}_i,\bm{\bar{V}}_j^-)+m \: \right |_{+},
\end{equation}
in which $m$ is the margin set as $0.1$ in our experiments, $d(\bm{x},\bm{y})$ is the distance between $\bm{x}$ and $\bm{y}$, and $|x|_+=x$ if $x>0$ and $0$ otherwise. For $d(\bm{\bar{u}},\bm{\bar{V}})$, a straightforward approach is calculating the distance between $\bm{\bar{u}}$ and weighted average of activity proposal features $Z(\bm{\bar{V}})$, but that will cause serious loss of structural information in activity proposals. As shown in~\cite{XuYS0S17}, point-to-subspace distance\footnote{https://en.wikipedia.org/wiki/Projection\_(linear\_algebra)\label{point}} is able to preserve the structural information and geometric property. 
In our problem, an image can be seen as a high-dimensional data point and video is a subspace spanned by activity proposals. Then the point-to-subspace distance is the Euclidean distance between an image point and its orthogonal projection on the subspace of videos. 

Considering that noisy proposals may mislead point-to-subspace distance, we use truncated bag of proposals in lieu of intact bag of proposals. To be exact, we denote truncated bag as  $\bm{\bar{V}}'=\bm{\bar{V}}[:,\mathcal{S}_b]$, in which $\mathcal{S}_b$ is the index set of proposals with top-$b$ GMIL weights $\hat{a}_i$. That means, we use the top-$b$ clean proposals in triplet loss.
With simple mathematical derivation\textsuperscript{\ref {point}},
the orthogonal projection of point $\bm{\bar{u}}$ on subspace $\bm{\bar{V}}'$ can be calculated as $\bm{\widetilde{V}}\bm{\bar{u}}$, where $\bm{\widetilde{V}}=\bm{\bar{V}}'((\bm{\bar{V}}')^{T}\bm{\bar{V}}')^{-1}(\bm{\bar{V}}')^{T}$.  
Then, the point-to-subspace distance between $\bm{\bar{u}}$ and $\bm{\bar{V}}'$, \emph{i.e.}, Euclidean distance between $\bm{\bar{u}}$ and $\bm{\widetilde{V}}\bm{\bar{u}}$, can be simplified as

\begin{eqnarray} \label{eqn:duv2}
\,\,d\left (\bm{\bar{u}},\bm{\bar{V}}' \right )
=&&\!\!\!\!\!\!\!\!\!\!\! \left \| \bm{\bar{u}}\!-\!\bm{\widetilde{V}}\bm{\bar{u}} \right \|_2^2\nonumber
\!=\!Tr((\bm{I}_{r}\!-\!\bm{\widetilde{V}})^{T}(\bm{I}_{r}\!-\!\bm{\widetilde{V}})\bm{\bar{u}}\bm{\bar{u}}^{T})\nonumber\\\
= &&\!\!\!\!\!\!\!\!\!\!\!\bm{\bar{u}}^{T}\bm{\bar{u}}-\left \langle \textbf{vec}(\bm{\widetilde{V}}),\textbf{vec}(\bm{\bar{u}}\bm{\bar{u}}^{T}) \right \rangle
\end{eqnarray}

By using $\widetilde{d}\left (\bm{\bar{u},\widetilde{V}} \right )$ to denote $\left \langle \textbf{vec}(\bm{\widetilde{V}}),\textbf{vec}(\bm{\bar{u}}\bm{\bar{u}}^{T}) \right \rangle$ and substituting (\ref{eqn:duv2}) into (\ref{eqn:stl_triplet}), we can arrive at

\begin{eqnarray} \label{eqn:new_triplet}
\!\!\!\!\!\!\!&&L_{triplet}=\sum_{i,j}\left |\: d(\bm{\bar{u}}_i,\bm{\bar{V}}_i^{'+})-d(\bm{\bar{u}}_i,\bm{\bar{V}}_j^{'-})+m \: \right |_{+}\nonumber\\
\!\!\!\!\!\!\!&&=\sum_{i,j}\left |\: \widetilde{d}\left (\bm{\bar{u}}_i,\bm{\widetilde{V}}_j^-\right )  \!-\! \widetilde{d}\left (\bm{\bar{u}}_i,\bm{\widetilde{V}}_i^+ \right )\!+\!m \: \right |_{+}.
\end{eqnarray}

Following~\cite{YaoMN15}, given an anchor sample $\bm{\bar{u}}_i$, we tend to select its hardest negative sample $\bm{\bar{V}}_{j}^{-}$ and the details are omitted here.
Based on (\ref{eqn:new_triplet}), we tend to minimize $L_{triplet}$ by optimizing GMIL module parameters $\bm{\theta}_{m}$ and projection module parameters $\bm{\theta}_{p}$. 

\subsubsection{Classification Loss with GMIL} \label{sec:cls_loss_MIL}
To ensure the training samples in each modality are semantically discriminative, we additionally use a semantic classifier to separate intra-modal training samples from different categories. To minimize the modality gap, we apply the same classifier for both images and videos. In particular, we add a softmax classification layer with model parameters $\bm{\theta}_c$ on top of the image features $\bm{\bar{u}}$ and the weighted average of proposal features $Z(\bm{\bar{V}})$. Given a mini-batch of video-image pairs $\{(\bar{\bm{V}_{i}},\bm{\bar{u}}_{i})|_{i=1}^n\}$ associated with one-hot label $\{\bm{y}_i|_{i=1}^n\}$, the classification loss is written as follows,
\begin{equation}\label{eq:class}
L_{class}=-\frac{1}{n}\sum_{i=1}^{n}\bm{y}_i^T (log(p(Z(\bm{\bar{V}}_{i})))+log(p(\bm{\bar{u}}_{i}))),
\end{equation}
in which $p(\cdot)$ denotes the prediction scores by using the softmax classification layer. Defining GMIL module parameters $\bm{\theta}_{m}=\{\bm{L},\bm{w}\}$, we tend to minimize $L_{class}$ by optimizing semantic classifier parameters $\bm{\theta}_{c}$, GMIL module parameters $\bm{\theta}_{m}$, and projection module parameters $\bm{\theta}_{p}$. 

\subsubsection{Adversarial Loss with GMIL}
To further minimize the modality gap across videos and images, adversarial learning~\cite{GoodfellowPMXWOCB14,WangYXHS17} is implemented as an interplay between discriminating modalities by learning a modality classifier and learning representations to confuse the modality classifier. 
In the process of discriminating modalities, we learn a modality classifier to discriminate the video modality from the image modality. The modality classifier is implemented as a binary classifier with model parameters $\bm{\theta}_d$, in which we assume the label of video (\emph{resp.}, image) modality is $1$ (\emph{resp.}, $0$).
In the process of learning representations to confuse the modality classifier, we expect the projected video/image features in the common feature space could fool the modality classifier. 
Considering that clean proposals have more representative feature distribution while the noisy proposals are scattered throughout the feature space, we apply the modality classifier on the weighted average of proposal features $Z(\bm{\bar{V}})$ for videos. Similar to the classification loss, the adversarial loss is formally defined as
\begin{equation}\label{eq:adv}
L_{adv} = -\frac{1}{n}\sum_{i=1}^{n} log(\delta (Z(\bm{\bar{V}}_{i}))+ log(1-\delta (\bm{\bar{u}}_{i})),
\end{equation}
where $\delta (\cdot)$ is the predicted probability of being from video modality.  
As adversarial learning is an interplay between discriminating modalities and learning representations, in the process of discriminating modalities, we tend to minimize $L_{adv}$ by optimizing the modality classifier parameters $\bm{\theta}_{d}$. On the contrary, in the process of learning representations, we tend to maximize $L_{adv}$ by optimizing projection module parameters $\bm{\theta}_{p}$ and GMIL module parameters $\bm{\theta}_{m}$.

\subsubsection{The Whole Algorithm}
We collect $L_{triplet}$, $L_{class}$, and $L_{adv}$ in (\ref{eqn:new_triplet}), (\ref{eq:class}), (\ref{eq:adv}) as the following total training loss:
\begin{equation} \label{eqn:total_loss}
L_{total} = \alpha \cdot L_{triplet} + \beta \cdot L_{class} -  L_{adv},
\end{equation}
where $\alpha$ and $\beta$ are trade-off parameters and empirically fixed as $0.1$ and $10$ respectively in our experiments.

Due to the adversarial loss $L_{adv}$ in (\ref{eqn:total_loss}), we play a minimax game by learning representations and discriminating modalities alternatingly.
By using $\bm{\theta}_{g}=\{\bm{\theta}_{p}, \bm{\theta}_{m}, \bm{\theta}_{c}\}$ to denote the model parameters in learning representations, our objective can be written as follows,
\begin{equation}
\underset{\bm{\theta}_{g}}{min} \,\underset{\bm{\theta} _{d}}{max}\, \  \alpha \cdot L_{triplet} + \beta \cdot L_{class} -  L_{adv},
\end{equation}
which can be optimized by updating $\bm{\theta}_{g}$ and $\bm{\theta}_{d}$ in an alternating manner. 
We leave the summary of our training algorithm to Supplementary due to space limitation.
In the testing, we pass the testing images and videos through our trained model, yielding 
the projected features $\bm{\bar{u}}$ (\emph{resp.}, $Z(\bm{\bar{V}})$) for images (\emph{resp.}, videos). Then, given a query image $\bm{u}_i$, we retrieve its relevant videos by ranking all the $l_2$ distances between $\bm{\bar{u}}_i$ and $Z(\bm{\bar{V}})$.






\section{Experiments}
In this section, we compare our APIVR approach with the state-of-the-art methods on three datasets and provide extensive ablation studies.

\subsection{Datasets Construction} \label{sec:exp_datasets}
To the best of our knowledge, there are no publicly available datasets of activity video-image pairs specifically designed for the AIVR task. Therefore, we construct video-image datasets for the AIVR task based on public video datasets, \emph{i.e.},  THUMOS'14\footnote{http://crcv.ucf.edu/THUMOS14/}, ActivityNet~\cite{HeilbronEGN15} and MED2017 Event\footnote{https://www.nist.gov/itl/iad/mig/med-2017-evaluation/} datatsets, in which THUMOS'14 and ActivityNet datasets are action-based datasets while MED2017 Event dataset is an event-based dataset. The difference between ``action" and ``event" lies in that an event generally consists of a sequence of interactive or stand-alone actions. 
The details of above three datasets are left to Supplementary. 
Based on the above three datasets, we aim to obtain activity images and activity video clips, which can be used to construct our datasets for AIVR task. 

To obtain activity video clips, considering that long videos may belong to multiple activity categories, we divide each long video into multiple short videos based on the activity temporal annotations to ensure that each short video only belongs to one activity category. Then, we sample a fixed number of consecutive key frames in each short video as a video clip. The number of key frames used in our experiments is $768$ for all datasets, which is large enough to cover at least one activity instance. 

To obtain activity images, we first locate the activity intervals in long videos according to activity temporal annotations. Then, we sample images from those activity intervals so that each image should belong to one activity category. 

With obtained activity images and activity video clips, we sample video clips and images from each category to form training pairs and testing pairs. Particularly, for THUMOS'14 dataset, we form 1500 training pairs and 406 testing pairs. For ActivityNet dataset, we form 4800 training pairs and 1200 testing pairs. For MED2017 Event dataset, we form 2200 training pairs and 404 testing pairs. 

\subsection{Implementation Details}

For images, we employ VGG model~\cite{SimonyanZ14a} to extract the fc7 layer features and then reduce the dimension from $4096$-dim to $128$-dim by PCA for the ease of memory and computation in our experiment.

For video clips, we use R-C3D model to generate activity proposals, which is pretrained on Sports-1M dataset and finetuned on UCF101 dataset~\cite{TranBFTP15}. We extract a $4096$-dim feature vector for each activity proposal and each video is represented by a bag of top-$60$ proposal features, \emph{i.e.}, $k=60$, by ranking the scores that may contain activities. In our geometry-aware triplet loss, we use top-$50$ proposals in each bag, \emph{i.e.}, $b=50$.

In the projection module, mapping functions $f_v(\cdot)$ (\emph{resp.}, $f_u(\cdot)$) are implemented as three fully-connected layers as follows. $f_v:\bm{V}(d_1\!=\!4096)\rightarrow 500\rightarrow 200\rightarrow \bm{\bar{V}}(r\!=\!64)$ and $f_u:\bm{u}(d_2\!=\!128)\rightarrow 100\rightarrow 80\rightarrow \bm{\bar{u}}(r\!=\!64)$. In our experiments, we use mAP@K, \emph{i.e.}, mean Aversion Precision based on top $K$ retrieved results, as the evaluation metric.

\begin{table}[tp]	
	\centering
	\resizebox{1.0\linewidth}{!}{
		\def\arraystretch{1.0}
		\begin{tabular}{|c|cccc|}
			\hline
			\multirow{2}{*}{\textbf{Method}}  &\multicolumn{4}{c|}{\textbf{mean Average Precision (mAP)}} \\ \cline{2-5}
			&\textbf{@10} &\textbf{@20} &\textbf{@50} &\textbf{@100}\\  \hline\hline
			APIVR (w/o $TL$) &0.3228&0.3096&0.2956&0.2875\cr\hline
			APIVR (w/o $AL$) &0.3278&0.3146&0.3026&0.2905\cr\hline
			APIVR (w/o $CL$) &0.2438&0.2389&0.2312&0.2306\cr\hline
			APIVR (w/o $GA$) &0.3531&0.3376&0.3204&0.3145\cr\hline
			APIVR (w/o $GMIL$) &0.3428&0.3368&0.3276&0.3102\cr\hline
			APIVR (w/o $Graph$) &0.3618&0.3521&0.3326&0.3285\cr\hline
			Full APIVR approach &{\bf 0.3812}&{\bf 0.3645}&{\bf 0.3459}&{\bf 0.3314}\cr
			\hline
	\end{tabular}}
	\caption{Comparision of our full APIVR approach and our special cases in terms of mAP@K on THUMOS'14. Best results are denoted in boldface.}
	\label{tab:1}
\end{table}

\begin{figure}
	\begin{center}
		\includegraphics[width=0.8\linewidth]{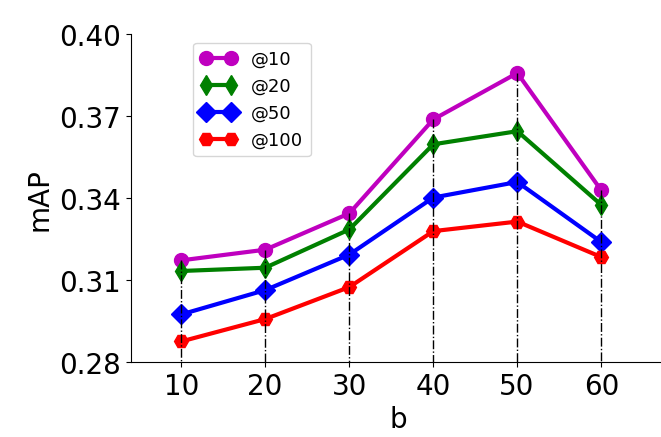}
	\end{center}
	\caption{The effect of top-$b$ proposals chosen from video bags to represent videos on THUMOS'14 dataset.}
	\label{fig:4}
\end{figure}

\begin{figure*}
	\begin{center}
		\includegraphics[width=0.9\linewidth]{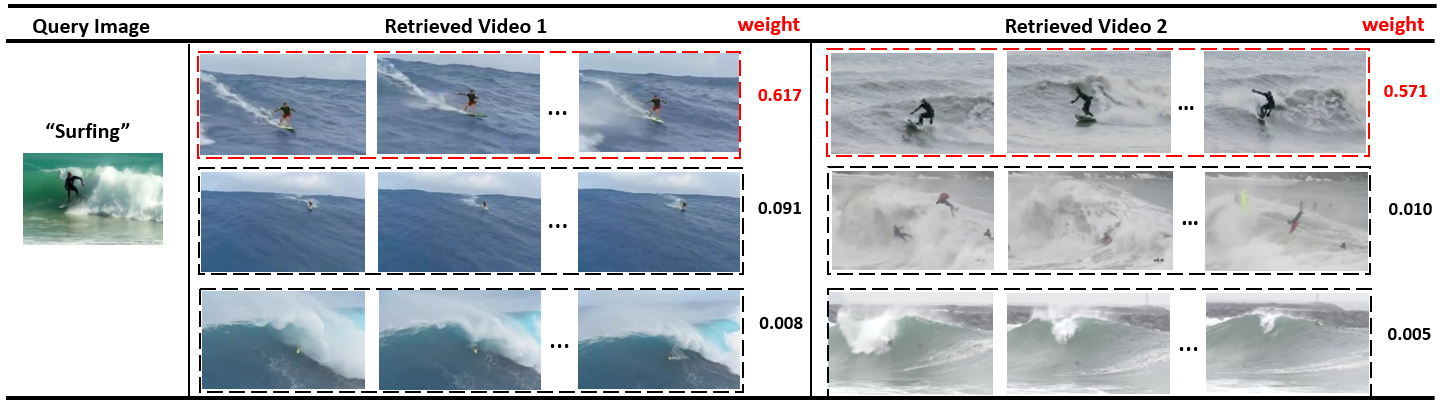}
	\end{center}
	\begin{center}
		\includegraphics[width=0.92\linewidth]{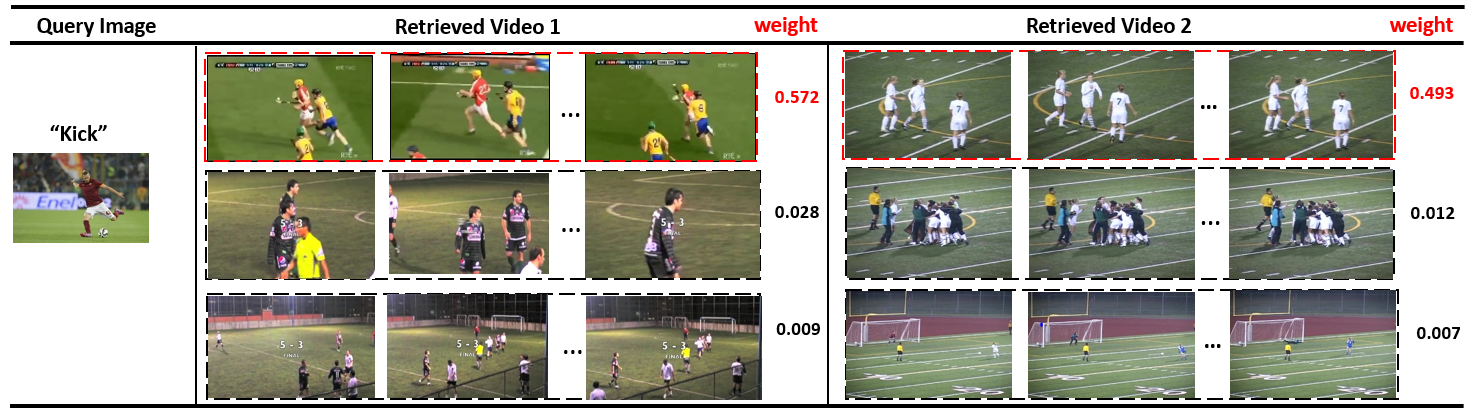}
	\end{center}
	\caption{Illustration of activity proposal weights learnt by our GMIL module on the ActivityNet dataset. The clean proposal is assigned with the highest weight (marked in red) and the other two noisy proposals are assigned with the lowest weights.}
	\label{fig:3}
\end{figure*}


\subsection{Ablation Studies} 
In order to explore the effectiveness of different components in our approach, we investigate some special cases of our approach. Specifically, we study the contributions of three types of losses by comparing with APIVR (w/o $TL$), APIVR (w/o $AL$), and APIVR (w/o $CL$), which are our three special cases by ablating Triplet Loss (TL), Adversarial Loss (AL), and Classification Loss (CL) respectively. Besides, to verify the benefit of geometry-aware triplet loss, we replace $d\left ( \bm{\bar{u}},\bm{\bar{V}}' \right )$ in (\ref{eqn:duv2}) with
$\left \| \bm{\bar{u}}-Z(\bm{\bar{V}}) \right \|_2^2$ and refer to this special case as APIVR (w/o $GA$). To demonstrate the effectiveness of our Graph Multi-Instance Learning (GMIL) module, we replace GMIL module in (\ref{eqn:a}) with MIL module in (\ref{eqn:mil1}), and name this case as APIVR (w/o $Graph$). Furthermore, we also assign uniform weights to proposals in each video instead of learning weights using GMIL module and name this special case as APIVR (w/o $GMIL$), which means that $Z(\bm{\bar{V}})=\frac{1}{k}\sum_{s=1}^{k}\bm{\bar{h}_{s}}$ in (\ref{eq:class}) (\ref{eq:adv}) and intact bags of activity proposals are used in (\ref{eqn:new_triplet}).

By taking THUMOS'14 dataset as an example, experimental results are reported in Table \ref{tab:1}. Obviously, we can see that APIVR (w/o $TL$), APIVR (w/o $AL$), and APIVR (w/o $CL$) are all inferior to our full APIVR approach, which indicates that each type of loss plays an essential role in our cross-modal framework and contributes to the overall performance. 
Based on the results of three losses, compared with adversarial loss and triplet loss, we can see that classification loss has more influence on the performance, which proves the significance of semantic classifier in our approach.
When using standard triplet loss instead of geometry-aware triplet loss, APIVR (w/o $GA$) suffers from a drop in performance, which demonstrates that it is beneficial to preserve the structural information and geometric property of activity proposals. 
Moreover, we can also note that the results of APIVR (w/o $GMIL$) are worse than the full APIVR approach, which proves the benefit of paying more attention to clean proposals based on our GMIL module. 
More results of ablating GMIL for each loss are provided in Supplementary. 
Finally, we can observe that APIVR (w/o $graph$) underperforms the full APIVR approach, which shows the advantage of inserting graph convolutional layer into MIL module. 

Recall that we use truncated bags of top-$b$ clean proposals in our geometry-aware triplet loss. To investigate the impact of $b$, we vary $b$ and report the performance of our full APIVR approach in
Figure \ref{fig:4}. 
We can observe that $b=50$ achieves the best performance, and the intact bags of proposals, \emph{i.e.}, $b=60$, may harm the performance because of the included noisy proposals.
When $b$ is very small (\emph{i.e.}, $b\leq 30$), too much useful information is discarded and thus the performance is also unsatisfactory. 

\begin{table*}[t]
	\centering
	\resizebox{0.95\linewidth}{!}{
		\def\arraystretch{0.92}	
		\begin{tabular}{|c|cccc|cccc|}
			\hline
			\multirow{2}{*}{\textbf{Methods}}&
			\multicolumn{4}{c|}{\textbf{THUMOS'14 dataset}}&\multicolumn{4}{c|}{\textbf{MED2017 Event dataset}}\cr\cline{2-9}
			&\textbf{mAP@10} &\textbf{mAP@20} &\textbf{mAP@50} &\textbf{mAP@100} &\textbf{mAP@10} &\textbf{mAP@20} &\textbf{mAP@50} &\textbf{mAP@100}\cr
			\hline\hline		
			ITQ&0.2613&0.2572&0.2477&0.2340&0.2284&0.2168&0.2127&0.2034\cr
			SpH&0.2131&0.2080&0.2033&0.1914&0.2044&0.1926&0.1878&0.1611\cr
			SKLSH&0.2004&0.1974&0.1951&0.1847&0.1956&0.1924&0.1883&0.1774\cr
			CBE-opt&0.2687&0.2601&0.2554&0.2483&0.2268&0.2128&0.2051&0.1984\cr
			\hline\hline
			MFH&0.2402&0.2398&0.2188&0.2128&0.2246&0.2192&0.2108&0.1994\cr
			SCM&0.2661&0.2576&0.2484&0.2395&0.2113&0.2041&0.1962&0.1924\cr
			CMFH&0.2545&0.2513&0.2466&0.2331&0.2262&0.2169&0.2101&0.2088\cr
			BPBC&0.2724&0.2706&0.2684&0.2571&0.2488&0.2501&0.2451&0.2402\cr\hline\hline
			JRL&0.2770&0.2656&0.2526&0.2411&0.2347&0.2278&0.2203&0.2198\cr
			CCL&0.3222&0.3188&0.3072&0.2949&0.2454&0.2417&0.2321&0.2267\cr
			JFSSL&0.2367&0.2351&0.2325&0.2241&0.2292&0.2218&0.2131&0.2064\cr
			Corr-AE&0.2266&0.2178&0.2096&0.2104&0.2032&0.2011&0.1971&0.1918\cr
			DSPE&0.2632&0.2544&0.2443&0.2312&0.2312&0.2246&0.2161&0.2004\cr
			CMDN&0.2927&0.2892&0.2754&0.2714&0.2328&0.2342&0.2250&0.2171\cr
			ACMR&0.3361&0.3274&0.3107&0.3061&0.2518&0.2401&0.2373&0.2244\cr
			DSCMR&0.3621&0.3523&0.3251&0.3188&0.2665&0.2576&0.2470&0.2381\cr
			\hline\hline
			{\bf APIVR}&{\bf 0.3812}&{\bf 0.3645}&{\bf 0.3459}&{\bf 0.3314}&{\bf 0.3049}&{\bf 0.2973}&{\bf 0.2867}&{\bf 0.2771}\cr
			\hline
	\end{tabular}}
	\caption{mAP@K of different methods on THUMOS'14 and MED2017 Event dataset. Best results are denoted in boldface.}
	\label{tab:3}
	
\end{table*}

\begin{table}[t]
	\centering
	\resizebox{0.9\linewidth}{!}{
		\def\arraystretch{0.9}
		\begin{tabular}{|c|cccc|}
			\hline
			\multirow{2}{*}{\textbf{Method}}  &\multicolumn{4}{c|}{\textbf{mean Average Precision (mAP)}} \\ \cline{2-5}
			&\textbf{@10} &\textbf{@20} &\textbf{@50} &\textbf{@100}\\  \hline\hline
			ITQ&0.1851&0.1704&0.1598&0.1414\cr
			SpH&0.1885&0.1843&0.1617&0.1551\cr
			SKLSH&0.1638&0.1595&0.1556&0.1474\cr
			CBE-opt&0.2044&0.1970&0.1842&0.1768\cr
			\hline\hline
			MFH&0.2155&0.2048&0.1977&0.1932\cr
			SCM&0.2285&0.2230&0.2166&0.2011\cr
			CMFH&0.2334&0.2318&0.2205&0.2155\cr
			BPBC&0.2352&0.2296&0.2184&0.2071\cr\hline\hline
			JRL&0.2266&0.2182&0.2177&0.2096\cr
			CCL&0.2358&0.2208&0.2138&0.2082\cr
			JFSSL&0.2166&0.2087&0.1958&0.1929\cr
			Corr-AE&0.2024&0.2012&0.1924&0.1866\cr
			DSPE&0.2212&0.2107&0.2079&0.2055\cr	
			CMDN&0.2422&0.2401&0.2288&0.2232\cr
			ACMR&0.2318&0.2224&0.2111&0.2091\cr
			DSCMR&0.2481&0.2344&0.2287&0.2122\cr
			\hline\hline		
			\textbf{APIVR}&{\bf 0.2635}&{\bf 0.2545}&{\bf 0.2488}&{\bf 0.2319}\cr
			\hline
	\end{tabular}}
	\caption{Performance of different methods in terms of mAP@K on the ActivityNet dataset. Best results are denoted in boldface.}
	\label{tab:4}
\end{table}

\subsection{ Visualization of Retrieved Videos} 
To better demonstrate the effectiveness of our GMIL module for identifying clean proposals, we provide two representative qualitative results in Figure \ref{fig:3}, in which the query image belongs to the category ``surfing" (\emph{resp.}, ``kick" ) in the top (\emph{resp.}, ``bottom" ) row.  We list top-2 retrieved videos for each query image. For each retrieved video, we show one proposal with the highest weight and another two proposals with the lowest weights. It is obvious that the proposals with the highest weight can capture the relevant activity while the other two proposals are less relevant or even background segments, which indicates the great advantages of our GMIL module in identifying clean proposals.

\subsection{ Comparisons with the State-of-the-art Methods}
We compared our APIVR approach with the state-of-the-art methods including single modality hashing methods CBE-opt~\cite{YuKGC14}, ITQ~\cite{GongL11}, SKLSH~\cite{RaginskyL09}, SpH~\cite{HeoLHCY12}, multiple modalities hashing methods MFH~\cite{YeLTXW17}, SCM~\cite{ZhangL14}, CMFH~\cite{DingGZ14}, BPBC~\cite{XuYS0S17}, and cross-modal retrieval methods Corr-AE~\cite{FengWL14}, CMDN~\cite{PengHQ16}, ACMR~\cite{WangYXHS17}, DSPE~\cite{WangLL16}, JRL~\cite{ZhaiPX14}, JFSSL~\cite{WangHWWT16}, CCL~\cite{PengQHY18}, DSCMR~\cite{Zhen_2019_CVPR}.
Among them, BPBC is a hashing method targeting at image-to-video retrieval task.
Although the method in \cite{AraujoG18} also targets at image-to-video retrieval, but it focuses on improving video Fisher Vectors using bloom filters and thus cannot be directly applied to our problem. 
Besides, Corr-AE, CMDN, ACMR, CCL, DSPE and DSCMR are deep learning-based methods and have achieved remarkable results in cross-modal retrieval task. For all baselines, we take the average of proposal features extracted by R-C3D as the video features and VGG fc7 features as the image features for fair comparison. The encoding length in the hashing methods is set to 128-bit. 

The experiment results are summarized in Table \ref{tab:3}, \ref{tab:4}. Compared with ACMR~\cite{WangYXHS17} method, which has a similar framework to ours, our superior performance confirms the advantages of preserving structural information using geometric projection and attending clean proposals using GMIL module. Obviously, we can see that our approach achieves significant improvement over all baselines in all scope of $K$ on both action-based and event-based datasets. 
For example, on the ActivityNet dataset with the largest number of categories, APIVR approach outperforms the other methods by about $2\%$ in all scope of $K$.

\section{Conclusion}
In this paper, we have proposed the first activity proposal-based image-to-video retrieval (APIVR) approach for the activity image-to-video retrieval task. We have incorporated graph multi-instance learning module into cross-modal retrieval framework to address the proposal noise issue, and also proposed geometry-aware triplet loss.
Experiments on three datasets have demonstrated the superiority of our approach compared to the state-of-the-art methods.

\section{Supplementary}
\subsection{Details of Datasets}
We construct our datasets based on three public datasets: THUMOS'14\footnote{http://crcv.ucf.edu/THUMOS14/}, ActivityNet~\cite{HeilbronEGN15} and MED2017 Event\footnote{https://www.nist.gov/itl/iad/mig/med-2017-evaluation/} datatsets, in which THUMOS'14 and ActivityNet datasets are action-based datasets while MED2017 Event dataset is an event-based dataset. The details of the above three datasets are introduced as follows:

\noindent\textbf{THUMOS'14 dataset: }The THUMOS'14 dataset consists of 2765 trimmed training videos and 200 untrimmed validation videos from 20 different sport activities. We merge similar categories such as ``cliff diving'' and ``diving'', and obtain a total number of 18 categories.

\noindent\textbf{ActivityNet dataset: }The ActivityNet dataset(1.3) contains 200 activity categories. Due to the limit of the GPU memory and speed, we only use the validation set with 4926 videos. Similar to THUMOS'14 dataset, we merge similar categories such as ``clean and jerk'' and ``snatch'', leading to in total 156 categories.

\noindent\textbf{MED2017 Event dataset: }ActivityNet dataset
The Multimedia Event Detection (MED) launched by TRECVID consists of more complicated events, which is also suitable to explore the AIVR task. We use the resources for the Pre-Specified Event portion of the MED2017 evaluation. The dataset has totally 200 trimmed videos distributed in 10 event categories.

\subsection{Training Algorithm}
Recall that our objective function is
\begin{equation} \label{eqn:total_object}
\underset{\bm{\theta}_{g}}{min} \,\underset{\bm{\theta} _{d}}{max}\, \  \alpha \cdot L_{triplet} + \beta \cdot L_{class} -  L_{adv},
\end{equation}
in which $\alpha$ and $\beta$ are trade-off parameters and empirically fixed as $10$ and $0.01$ respectively. The problem in (\ref{eqn:total_object}) is a minimax problem, which can be optimized by updating $\bm{\theta}_{g}$ and $\bm{\theta}_{d}$ in an alternating manner. The details of our training process are shown in Algorithm \ref{alg:1}, in which we set the number of image-video pairs in a mini-batch $n=64$, the number of generation steps $t=50$, and the learning rate $\lambda=0.0001$.

\begin{algorithm}
	\begin{algorithmic}[1]
		\renewcommand{\algorithmicrequire}{\textbf{Input:}}
		\renewcommand{\algorithmicensure}{\textbf{update until convergence:}}
		\REQUIRE Mini-batches of video-image pairs $\{(\bm{V}_{i}, \bm{u}_{i})|_{i=1}^n\}$ and the associated labels $\{\bm{y}_{i}|_{i=1}^n\}$. The number of steps $t$, learning rate $\lambda$, and trade-off parameters $\alpha, \beta$.
		\ENSURE	 
		\FOR{t steps}
		\STATE update $\bm{\theta}_{g}$ by \textbf{descending} stochastic gradients:
		
		$\bm{\theta}_{g}\leftarrow \bm{\theta}_{g}-\lambda \cdot \bigtriangledown_{\bm{\theta}_{g}} (\alpha \cdot L_{triplet}+\beta \cdot L_{class}- L_{adv}).$
		
		\ENDFOR
		\STATE update parameter $\bm{\theta}_{d}$ by \textbf{ascending} stochastic gradients:
		
		$\bm{\theta}_{d}\leftarrow \bm{\theta}_{d}+\lambda  \cdot \bigtriangledown_{\bm{\bm{\theta}}_{d}}(\alpha \cdot L_{triplet}+\beta \cdot L_{class}- L_{adv})$.
		\RETURN{Model parameters $\bm{\theta}_{g}$ and  $\bm{\theta}_{d}$}
	\end{algorithmic}
	\caption{The training process of our APIVR approach.}
	\label{alg:1}
\end{algorithm}

\begin{small}
	\bibliographystyle{aaai}
	\bibliography{egbib}
\end{small}

\end{document}